\documentclass[11pt,a4paper]{article}
\usepackage[hyperref]{acl2021}
\usepackage{times}
\usepackage{latexsym}

\usepackage{microtype}
\usepackage{graphicx}
\usepackage{linguex}
\usepackage{booktabs}
\usepackage{xcolor}

\aclfinalcopy 

\title{A Targeted Assessment of Incremental Processing in Neural Language Models and Humans}

\author{Ethan Gotlieb Wilcox \\
  Harvard University \\ Department of Linguistics \\
  \texttt{wilcoxeg@g.harvard.edu} \\\And
  Pranali Vani \\
  MIT \\ Brain and Cognitive Science \\
  \texttt{pvani@mit.edu} \\\And
  Roger P.\ Levy \\
  MIT \\ Brain and Cognitive Science \\
  \texttt{rplevy@mit.edu} \\ }

\date{}

\begin{document}

\setlength{\Exlabelwidth}{0.7em}
\setlength{\Exlabelsep}{0.7em}
\setlength{\SubExleftmargin}{1.3em}
\setlength{\Extopsep}{4pt}

\maketitle

\begin{abstract}

We present a targeted, scaled-up comparison of incremental processing in humans and neural language models by collecting by-word reaction time data for sixteen different syntactic test suites across a range of structural phenomena. Human reaction time data comes from a novel online experimental paradigm called the \textit{Interpolated Maze} task. We compare human reaction times to by-word probabilities for four contemporary language models, with different architectures and trained on a range of data set sizes. We find that across many phenomena, both humans and language models show increased processing difficulty in ungrammatical sentence regions with human and model `accuracy' scores (\`a la \citet{marvin2018targeted}) about equal. However, although language model outputs match humans in direction, we show that models systematically under-predict the difference in magnitude of incremental processing difficulty between grammatical and ungrammatical sentences. Specifically, when models encounter syntactic violations they fail to accurately predict the longer reaction times observed in the human data. These results call into question whether contemporary language models are approaching human-like performance for sensitivity to syntactic violations.

\end{abstract}

\section{Introduction}

A substantial body of work has investigated contemporary language models (LMs) by assessing whether their behavior is consistent with the rules of syntax \citep{hu2020systematic, marvin2018targeted, warstadt2020blimp}.\footnote{Data and code for this paper can be found online at \url{https://github.com/wilcoxeg/targeted-assessment-imaze}} Among other structures, these studies have investigated agreement \citep{linzen2016assessing, gulordava2018colorless} long distance dependencies \citep{wilcox2018rnn}, pronominal and particle licensing \citep{jumelet2018language, futrell2019neural}, and expectations for phrase-level constituents \cite{futrell2018rnns}. Many of the studies which report aggregate behavior across a broad number of phenomena focus on accuracy scores, or the proportion of time LMs or human subjects in an online experiment prefer a grammatical variant in matching grammatical / ungrammatical sentence pairs. While these investigations provide much insight, they collapse a crucial dimension of comparison, namely the difference in magnitude between the grammatical and ungrammatical conditions. As long as the direction of their predictions are the same, an LM which finds grammatical conditions only marginally worse than their corresponding ungrammatical counterpart will receive the same score as a model that displays large differences between the two conditions.

At the same time, a related line of work has investigated the quantitative relationship between incremental predictions of language models and human reaction times \cite{hale2001probabilistic, levy2008expectation}. \citet{smith2013effect} found that this relationship is log-linear across multiple orders of magnitude for $3$-gram models, and recent investigations have shown that this holds for contemporary neural network models as well \cite{wilcox2020predictive, goodkind2018predictive}. So far, this work has largely focused on the aggregate relationship, instead of isolating individual phenomena in targeted testing environments.

\begin{table*}[!ht]
    \centering
    \footnotesize
    \begin{tabular}{llp{7cm}}
        \toprule
        Test Suite Name & Tag & Example \\
        \midrule
        Wh-Cleft Structures & Cleft & What she \textbf{did/spied} was \underline{\textbf{see the giraffe/the giraffe}}  \\
        \midrule
        Filler-Gap Dependency, Subject Gap & FGD-subj & I know \textbf{who/that} \textbf{\_\_\_/\underline{my mom}} \underline{sent} the gift to Kim. \\
        Filler-Gap Dependency, Object Gap & FGD-obj &  I know \textbf{who/that} my mom sent \textbf{\_\_\_/\underline{the gift}} \underline{to} Kim. \\
        Filler-Gap Dependency, PP Gap & FGD-pp & I know \textbf{who/that} my mom sent the gift to \textbf{\_\_\_/\underline{Kim}} \underline{yesterday}.  \\
        \midrule
        Main Verb/Reduced RC Gardenpath & MVRR & The ship \textbf{$\emptyset$/that was} \textbf{sunk/steered} in the storm \underline{had gold}. \\
        \midrule
        NPI Licensing, \textit{any}, Subj RC Modifier & NPL-any-src & \textbf{No/The} senator that \textbf{no/the} man likes has won \underline{any} votes.  \\
        NPI Licensing, \textit{any}, Obj RC Modifier & NPL-any-orc & \textbf{No/The} senator that likes \textbf{no/the} man has won \underline{any} votes. \\
        NPI Licensing, \textit{ever}, Subj RC Modifier & NPL-ever-src & \textbf{No/The} senator that \textbf{no/the} man likes has \underline{ever} won.  \\
        NPI Licensing, \textit{ever}, Obj RC Modifier & NPL-ever-orc & \textbf{No/The} senator that likes \textbf{no/the} man has \underline{ever} won. \\
        \midrule
        Subject-Verb Number Agr., Subj RC Modifier & SVNA-src & The \textbf{lawyer/lawyers} that helped the mayor \underline{\textbf{is/are}} tall. \\
        Subject-Verb Number Agr., Obj RC Modifier& SVNA-orc & The \textbf{lawyer/lawyers} that the mayor hired \underline{\textbf{is/are}} very tall. \\
        Subject-Verb Number Agr., PP Modifier& SVNA-pp & The \textbf{lawyer/lawyers} next to the mayor \underline{\textbf{is/are}} very tall. \\
        \midrule
        Reflexive Anaphora, Masc., Subj RC Modifier & RNA-m-src  & The \textbf{dukes/duke} that hunted deer saw \underline{\textbf{himself/themselves}} in the mirror.  \\
        Reflexive Anaphora, Masc., Obj RC Modifier & RNA-m-orc  & The \textbf{dukes/duke} that the men hate saw \underline{\textbf{himself/themselves}} in the mirror. \\
        Reflexive Anaphora, Fem., Subj RC Modifier & RNA-f-src & The \textbf{queens/queen} that hunted deer saw \underline{\textbf{herself/themselves}} in the mirror.  \\
        Reflexive Anaphora, Fem., Obj RC Modifier & RNA-f-orc & The \textbf{queens/queen} that the men hate saw \underline{\textbf{herself/themselves}} in the mirror.  \\
        \bottomrule
    \end{tabular}
    \caption{The sixteen test suites evaluated in this paper. Sentence regions which are manipulated to form the four conditions in each test suite are indicated with bold. Critical regions are underlined.}
    \label{tab:test_suites}
    \vspace{-0.5cm}
\end{table*}

We combine these two approaches with a targeted assessment of incremental processing in neural language models and humans. We collect incremental processing data on a series of sixteen test suites, adapted from \citet{hu2020systematic}, each of which targets a different syntactic phenomenon. For LM incremental processing data, we collect by-word probabilities for four contemporary neural network architectures. For human incremental processing data, we use by-word reaction times (RTs). We collect these by deploying a novel online measurement paradigm called the \textit{Interpolated Maze}, which is based on the Maze task \cite{forster2009maze}. In the Maze task, participants must read a sentence incrementally by selecting the correct word from two possible continuations, one of which is ungrammatical. The time it takes participants to select the correct choice has been shown to effectively capture incremental processing cost and can be deployed at scale \cite{boyce2020maze}.

We deploy three analysis techniques to investigate how well models capture the human incremental processing data. First, we compute accuracy metrics (for LMs) and consistency scores (for humans) for each of our test suites, which correspond to the proportion of the time behavior is consistent with the relevant grammatical rules. We find that, for this analysis, humans and machine performance is about equal. Next, we compare the observed reaction-time slowdown between grammatical/ungrammatical conditions within a test suite to the slowdown predicted by each of our models. For this analysis we use the methodology developed by \citet{van2018modeling}, who use a \textit{ms/bit} (milliseconds of reaction time per bit of surprisal) conversion metric derived from a fitted regression model to convert between the outputs of LMs and slowdowns in human reaction times. We find that models systematically under-predict the observed human data. In our third analysis, we train a linear regression models to predict reaction times from probabilities in non-critical sentence regions, and show that these models are relatively poor at predicting reaction times in critical sentence regions. That is, in areas of the sentence where human reaction time is influenced by grammatical violations, LM probabilities routinely under-predict human processing difficulty as measured by reaction time. Taken together, these results indicate that contemporary neural network languages models are systematically less sensitive to grammatical violations compared to humans.

\section{Methods}

We collect incremental processing data on a series of test suites, each of which targets an individual syntactic phenomenon. Composition of the test suites is described in Section \ref{sec:test_suites}. Methods used to collect incremental processing data are outlined in Section \ref{sec:maze}, for human reaction times. Section \ref{sec:models} describes the models tested. Linear Regression Models used to predict reaction times from model outputs will be referred to as `Linear Fits' to avoid confusion with Language Models. 

\subsection{Syntactic Test Suites} \label{sec:test_suites}

We use sixteen test suites for syntactic generalization, adapted from \citet{hu2020systematic}. Test suites consist of 20-25 items. Each item appears in four conditions, two grammatical and two ungrammatical.\footnote{For the MVRR test suites, the `ungrammatical' conditions are plausibly licensed by the grammar, but are unlikely. Following convention in linguistics, ungrammatical sentences will be marked with a *. } Table \ref{tab:test_suites} gives the name of each test suite, an example, as well as a tag, which we will use to refer to that suite in figures. When test suites have modifiers they always included distractors of the opposite grammatical category. For example singular reflexive anaphora sentences with subject relative clause modifiers would have a plural noun in the relative clause (e.g.  \textit{The bishop who likes the kings saw *themselves/himself in the mirror.}) 

Following the logic from \citet{hu2020systematic}, each test suite comes with two or more criteria, which specifies an inequality that should hold in a particular \textit{critical region} if model behavior follows the rules of the relevant grammatical construction. Accuracy scores for each test suite are generated by computing the proportion of the time the inequality holds within the critical region, across items in a test suite. In \citeauthor{hu2020systematic}, test suites include criteria that correspond to 2-way contrasts between grammatical/ungrammatical conditions as well as 2x2 interactions between four conditions. We only look at the 2-way contrasts, here. 

The incremental processing measure we derive from a language model to determine its accuracy according to a suite's inequality predictions is \textit{surprisal}. Surprisal is the inverse log probability of a word given its context: $S(x_i) = -\log_2 p(x_i|x_1...x_{i-1})$, measured in bits. In this paper, we novelly extend the usage of these inequalities to determine a human \textit{consistency score} for each test suite, by checking the mean reaction times for the various conditions of each item in the suite against the suite's criteria. For naturalistic corpus materials, the effect of surprisal on human reaction times has been shown to be linear \cite{smith2013effect,goodkind2018predictive,wilcox2020predictive}, motivating this usage of syntactic generalization criteria on human reading patterns.  We use the same criteria as described in Appendix B of \citet{hu2020systematic}.

\begin{figure*}[t]
    \centering
    \includegraphics[width=0.95\textwidth]{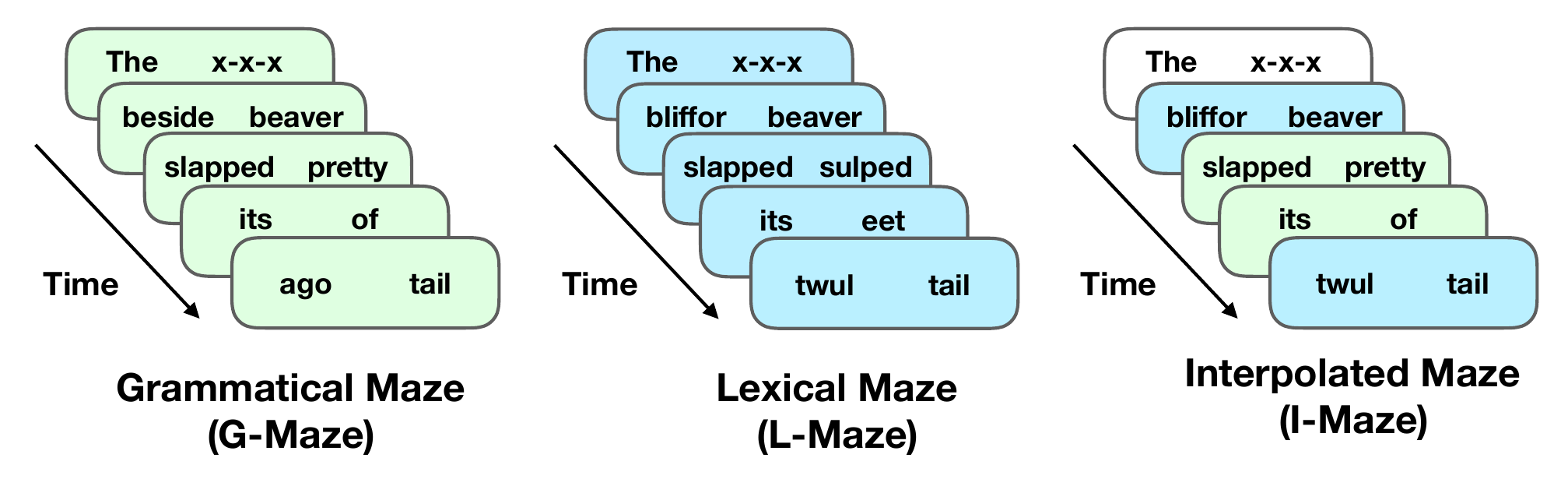}
    \caption{The Maze Task: Participants read the sentence word-by-word. At each index they must select the correct continuation. For this study, we introduce the \textit{Interpolated Maze}, which is a blend of G-Maze and L-Maze. }
    \label{fig:maze}
\end{figure*}

To walk through a single test suite in detail, \ref{ex:mvrr} gives an example of all four conditions of the \textit{Main Verb / Reduced Relative Clause} suite, with critical regions underlined.

\ex. \label{ex:mvrr}
\small
\a. The artist drawn a portrait \underline{was impressed} with the work. {\sc [reduced, unambiguous]}
\b. The artist that was drawn a portrait \underline{was impressed} with the work. {\sc [un-reduced, unambiguous]}
\c. The artist painted a portrait \underline{was impressed} with the work. {\sc [reduced, ambiguous]}
\d. The artist that was painted a portrait \underline{was impressed} with the work. {\sc [un-reduced, ambiguous]}

The logic of the test suite relies on the fact that strings like \textit{painted} are ambiguous between active past-tense main verbs and passive participles that introduce a reduced relative clause. On the other hand, verbs like \textit{drawn} unambiguously introduce a reduced relative clause. If subjects believe that the ambiguous form of the verb introduces a main verb, they should find the critical-region verb \textit{was impressed} surprising. That is, relative to the {\sc [reduced, ambiguous] } conditions, not reducing the verb or using an unambiguous verb should make the critical region less surprising (\ref{pred:1} and \ref{pred:2} below). Furthermore, the effect of not reducing the relative clause should be smaller for unambiguous verbs than for ambiguous ones (\ref{pred:3}).

If we denote for convenience $S_x(w_i)$ as the surprisal of word $w_i$ in the context of version $x$ of a test suite item, then the following list outlines these three predictions as inequalities, which we used to determine accuracy scores on our test suites.

{\footnotesize
\begin{enumerate}
    \item $S_c$(was impressed) $>$ $S_d$(was impressed) \label{pred:1} \vspace{-0.3cm}
    \item $S_c$(was impressed) $>$ $S_a$(was impressed)\label{pred:2}  \vspace{-0.3cm}
    \item ($S_c$(was impressed) - $S_d$(was impressed)) $>$ ($S_a$(was impressed) - S$_b$(was impressed)) \label{pred:3}  \vspace{-0.15cm}
\end{enumerate}
}

To foreshadow our results, the \textbf{MVRR} panels of Figures~\ref{fig:effect_results} and in Appendix A show that all three of these criteria are met for most items both by all models and by human average reaction times. Unlike our other test suites, these predictions do not correspond to contrasts between sentences that vary based on their grammaticality, but rather on predictive processing that prefers the main-verb analysis for locally ambiguous strings.


\subsection{The Interpolated Maze Task} \label{sec:maze}

Human reaction time data was collected via a novel implementation of the Maze Task \cite{forster2009maze} which we call the \textit{Interpolated Maze}. In a maze task participants read through a sentence; at each index they are presented with two possible continuations, one word is a plausible next-word in the sentence and the other word is a distractor. Participants must select the correct continuation by pressing a key on their keyboard. Figure \ref{fig:maze} shows a cartoon of this process for three variants of the Maze Task. In the G(rammatical)-Maze version, the distractor word is a word of English, only it does not constitute a grammatical continuation. In the L(exical)-Maze variant, the word is a non-English nonce word. If participants select the wrong continuation, the trial ends and they begin reading the next sentence. The time it takes participants to select the correct word by pressing a key has been shown to be a robust measure of incremental processing difficulty, with slowdowns occurring on target words instead of in subsequent spillover regions as is the case with other online processing measures such as self-paced reading \citep{boyce2020maze}. 

Of these two variants, G-Maze has been shown to produce higher sensitivity results than L-Maze \cite{boyce2020maze}, however because each index must present one possible continuation, it cannot be used be used for items that have ungrammatical conditions. At the critical choice point, both the distractor and the continuation would be ungrammatical and participants would not know which continuation to select. To solve this problem we deploy a novel variant of the maze task called \textit{Interpolated Maze}, or I-Maze. In I-Maze, we interweave G-Maze and L-Maze choices, with L-Maze distractors in critical regions where one of the conditions is ungrammatical. Participants are instructed to choose English words over nonce-words, thus making the `right' choice in these regions unambiguous. In order not to clump L-Maze distractors only in critical regions, we randomly sample $\sim$25\% of all other words and render them as L-Maze choices. For a full comparison of I-Maze, G-Maze and L-Maze see \citet{vani2021maze}. G-Maze distractors were generated with the scripts provided in \citet{boyce2020maze}, which uses a neural-network based language model to automatically generate high surprisal distractor words. Nonce words were generated with Wuggy \cite{keuleers2010wuggy}. Experiments were hosted on Ibex Farm \cite{drummond2013ibex}, with participants recruited on Amazon M-Turk. reaction time data for each item was collected from thirty separate participants.

\begin{figure*}
    \centering
    \includegraphics[width=0.99\textwidth]{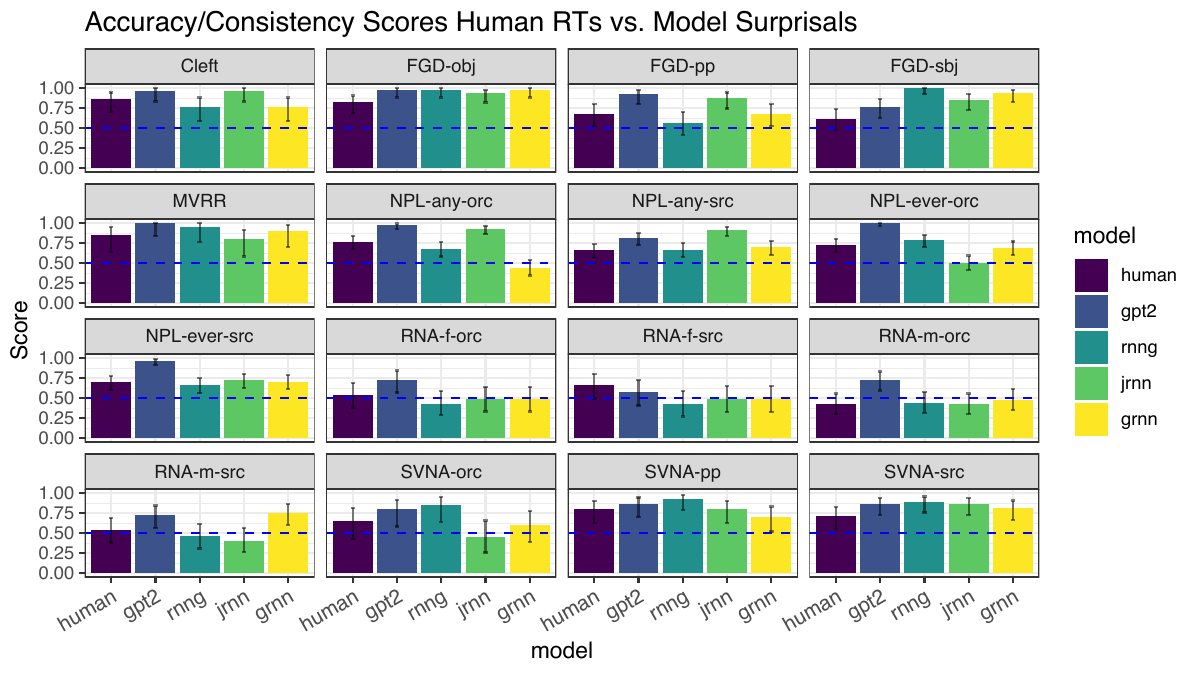}
    \caption{Comparison between human consistency scores and model accuracy scores. Averages are taken across all predictions within a test suite, error bars are 95\% binomial confidence intervals. Scores are similar between humans and models}
    \label{fig:acc_results}
\end{figure*}

\subsection{Models Tested} \label{sec:models}

\noindent \textbf{JRNN} is the `BIG LSTM+CNN Inputs' from \citet{jozefowicz2016exploring}. It was trained on the One Billion Word Benchmark \citep{chelba2013one} with two hidden layers of 8196 units each and CNN character embeddings as input.

\noindent \textbf{GRNN} is the best-performing model described in the supplementary materials of \citet{gulordava2018colorless}. It was trained on 90 million tokens of English Wikipedia with two hidden layers of 650 hidden units.

\noindent \textbf{GPT-2} is the model presented in \citet{radford2019language}, and was trained on ~40GB of internet text. We use the version of GPT-2 available through the \texttt{Language Modeling Zoo} distribution\footnote{\url{https://cpllab.github.io/lm-zoo/index.html\#welcome-to-lm-zoo}}

\noindent \textbf{RNNG} \cite{dyer2016rnng} jointly models a sentence as well as its syntactic parse. The model explicitly represents parse trees and composes partially built phrase structures. Models are supervised with Penn-Treebank style parses during training. We use the average of the three RNNG-BLLIP-LG models from \citet{hu2020systematic}.

\subsection{Addressing Two Possible Confounds}

Before we turn to our results, we will briefly address two possible confounds with our methods: First, while it may be the case that the relationship between surprisal and reaction time is linear in most sentence areas, this linearity may break down in high surprisal regions regardless of the underlying grammaticality of the sentence. Thus, any potential badness of our linear fits in critical regions is an epiphenomenon of the fact that they were trained in regions where the linearity holds and tested in regions where it does not. While there is some evidence that the linear relationship between surprisal may flatten off in high surprisal regions for self-paced reading (see, e.g. Figure 1 in \citet{wilcox2020predictive}), data collected for Maze task for both GRNN and a large Transformer model shows that the linear relationship holds even in very high surprisal regions, even exceeding 20 bits \citep{boyce2020amaze} (see, especially Figure 3).

The second confound has to do with the Interpolated Maze task. It may be the case that switching between tasks incurs a cognitive load, thus ungrammatical sentence regions might be read more slowly, but only because they are always associated with a switch from grammatical to lexical distractors. This could be worrisome, however we find that reaction times in non-critical regions for L-Maze decisions are actually slightly \textit{faster} than G-Maze decisions ($p<0.001$ by a $t$-test). Furthermore, all of our reported contrasts are between L-Maze items, so this is controlled for in our analyses.

\begin{figure*}
    \centering
    \includegraphics[width=\textwidth]{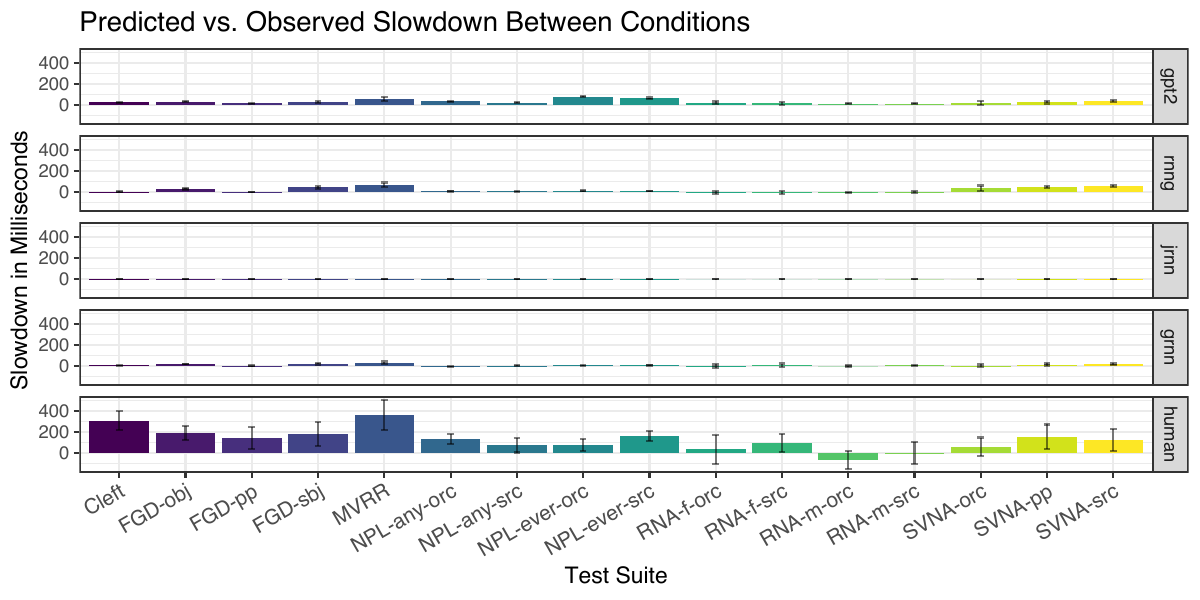}
    \caption{Comparison between human and (predicted) model reaction-time slowdows between grammatical and ungrammatical conditions. Averages are taken across all predictions within a test suite, error bars are 95\% confidence intervals. Models systematically under-predict the observed slowdown.}
    \label{fig:effect_results}
\end{figure*}

\section{Results}

\subsection{Test Suite Accuracy} \label{sec:accuracy}

In this section we discuss test suite accuracy scores, which are computed using the predictions associated with each test suite. For models, success on a prediction means that the model found material in a specified \textit{critical region} more probable in the grammatical condition than the ungrammatical condition. For humans, a corresponding metric, \textit{consistency scores}, report the proportion of times the critical region material was read more quickly in the grammatical condition than in the ungrammatical condition. Scores are calculated across the total number of items in a test suite. Because multiple subjects provided reaction time data for each item, we first average item-level data across all participants before calculating consistency scores.

The accuracy/consistency scores for each of our test suites can be seen in Figure \ref{fig:acc_results}. In this figure each facet represents the results from a single test suite, which aggregates across two or more predictions. A full breakdown of test suite by prediction can be seen in Appendix B. Chance, which is 50\% accuracy, is marked with a dashed blue line.

Humans perform above chance on 13/16 test suites. Human RTs are at or below chance for 3/4 of the Reflexive Anaphora agreement tests and the Subject-Verb Number Agreement with an Object Relative Clause modifier. For the Reflexive Anaphora tests, the low scores are driven by poor performance when the noun that must be matched is singular, such as in \textit{The lawyer who the judges fear hurt herself/*themselves}. Notably, human reaction times for negative polarity items and for number agreement on verbs and reflexive pronouns are known to be susceptible to facilitatory interference effects from intervening attractors of the sort that are used in our test suites \citep{vasishth-etal:2008,jager-etal:2020-interference-patterns}. In general, human consistency scores in this study are below that reported in \citet{marvin2018targeted}, who use an offline forced-choice paradigm, in which participants must judge which of two sentences sounds more natural. Nevertheless, for the vast majority of test suites, humans show robust sensitivity to the grammatical effects being tested, and failure is due to specific biases, such as the singular reflexive behavior discussed above, not general insensitivity to the manipulations.

Table \ref{tab:accuracy_correlations} shows the cross-suite correlations between human consistency scores and model accuracy scores. The relatively strong correlation scores indicate that the strength of signal for a syntactic generalization in model surprisal differentials is predictive of the signal-to-noise ratio for the generalization in human reaction times.

\begin{table}
    \centering
    \begin{center}
    \begin{tabular}{ccc}
    \toprule
        Model & Correlation & $p$-value  \\
        GRNN & 0.45 & $0.07$ \\
        JRNN & 0.68 & $<0.01$ \\
        GPT2 & 0.71 & $<0.01$ \\
        RNNG & 0.65 & $<0.01$ \\
        \bottomrule
    \end{tabular}
    \end{center}
    \caption{Correlations between model accuracy scores and human consistency scores across test suites.}
\label{tab:accuracy_correlations}
\end{table}

\begin{figure*}
    \centering
    \includegraphics[width = 0.31\textwidth]{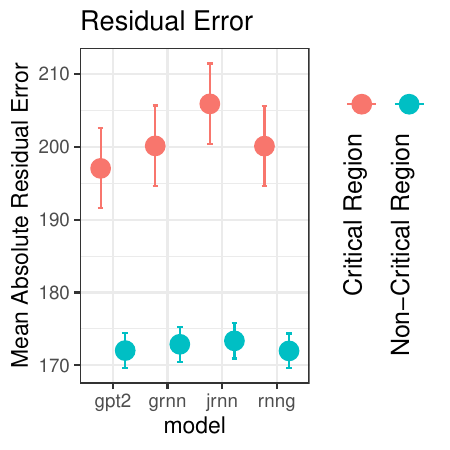}
    \hspace{-0.5cm}
    \includegraphics[width = 0.4\textwidth]{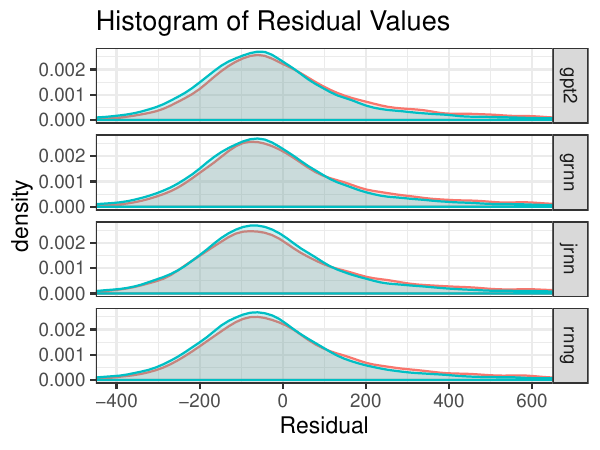}
    \includegraphics[width = 0.29\textwidth]{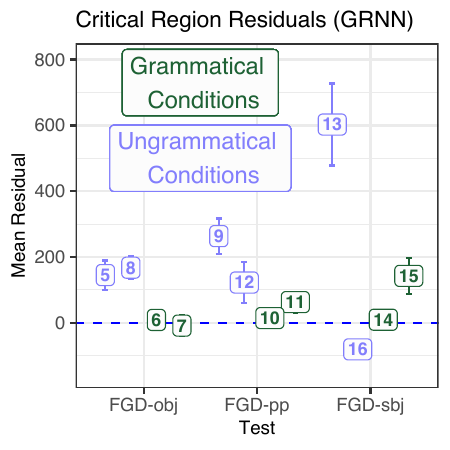}
    \caption{Residuals for reaction times in critical regions from a linear fit trained to predict reaction times from surprisal values in non-critical sentence regions. The left facet shows mean absolute residual error and the center shows a histogram of the raw values, with larger residuals in critical regions. The right facet shows a breakdown by condition for the Filler--Gap Dependency tests (GRNN model), with larger residual values in the ungrammatical conditions. For this plot, labels indicate condition name, with a reference provided in Appendix A. Error bars are 95\% confidence intervals.}
    \label{fig:resid_analysis}
\end{figure*} 

\subsection{Slowdown Between Conditions} \label{sec:slowdown}

In this section we turn to the size of the contrast between grammatical and ungrammatical conditions. For humans, this contrast indicates a slowdown, where critical regions of ungrammatical sentences are read more difficultly than their corresponding grammatical variants. For LMs, this contrast indicates a surprisal difference, where ungrammatical conditions are more surprising than their grammatical counterparts. Do differences in surprisal accurately predict the slowdowns observed in human reaction time data?

To derive a predicted reaction-time slowdown from the model surprisals, we followed the methodology outlined in \citet{van2018modeling}. This approach draws on the fact that the relationship between surprisal and human reaction time is linear across multiple orders of magnitude \cite{smith2013effect, wilcox2020predictive}, including for Maze data \cite{boyce2020amaze}. For each LM, we trained a linear fit that predicts reaction time from surprisal value at the word-level. The model is fit on RTs from all L-Maze distractor trials, critical and non-critical region alike, and includes word frequency and word length as additional predictors, with random slopes for each item and each participant. The linear model's surprisal estimate, therefore, is the slowdown in processing time predicted for each bit of surprisal. We treat this number as a scalar and multiply it by the difference in surprisal between conditions to derive the total predicted slowdown due to syntactic violation from the language models. For all of our fits, we found a significant effect for all of our predictors. The estimates for each model's surprisal term are given in Table \ref{tab:surp_estimates}.

\begin{table}
\centering
\begin{tabular}{ccc}
    \toprule
    Model & Surprisal Estimate & $p$-value  \\
    GRNN & $8.8ms/bit$ & $<0.001$ \\
    JRNN & $0.5ms/bit$ & $<0.05$ \\
    GPT2 & $12.0ms/bit$ & $<0.001$ \\
    RNNG & $19.0ms/bit$ & $<0.001$ \\
    \bottomrule
    \end{tabular}
    \vspace{-0.3cm}
    \caption{Surprisal Estimates from Linear Fits}
\label{tab:surp_estimates}
\end{table}

The results from this analysis can be seen in Figure \ref{fig:effect_results}, with the various test suites on the x-axis and observed or predicted slowdowns on the y-axis. As with accuracy scores, we average across predictions within each test suite. Humans demonstrate positive slowdowns in 11/16 test suites, with reflexive anaphora again proving the exception to the general trend. As is evident from the height of the bars, models systematically under-predict the slowdown observed in the human data. Models' predictions are outside of the 95\% confidence intervals for the humans slowdowns in 7/16 test suites for GPT2, 8/16 for RNNG, 9/16 for GRNN and 12/16 for JRNN. The mean predicted difference between models and humans across all test suites is 95$ms$ (GPT2), 107$ms$ (RNNG), 117$ms$ (GRNN) and 126$ms$ (JRNN). These data indicate that models are less sensitive to the contrast between grammatical and ungrammatical conditions than are humans, at least in this controlled testing environment.

\subsection{Residuals} \label{sec:residuals}


In this section, we discuss a follow-up analysis conducted to validate the conclusion that models are under-predicting reaction times in critical regions. To do this, we train linear fits on data from the non-critical regions, and get their residuals on data from these regions as well the critical regions. The linear fits are exactly the same as the ones described in the previous section, except instead of being trained on both critical and non-critical L-Maze trials, they are trained on non-critical L-Maze trials alone. If the conclusion from the last section is correct, then we should see larger residuals for the critical-region data then for the non-critical region data.

The results from this analysis can be seen in the right and center facets of Figure \ref{fig:resid_analysis}. The left facet shows the mean absolute value of the residuals for each of our LMs, both for the critical and non-critical region. The center facet shows a histogram of the same data. From both plots it is clear that the critical region residuals are greater than the residuals computed for words in other regions of the sentence. From the histograms, we can see that the critical region residuals are systematically higher on average than the non-critical region residuals. This indicates that the models under-predict the RT values in the critical regions. 

The difference between residuals provides additional evidence that models under-predict reaction times in critical regions compared to words in other parts of the sentence. However, it does not show that models under predict reaction times specifically for \textit{ungrammatical} sentences. To investigate this, we break down average residual by condition, within each of our sixteen test suites. The full results for this breakdown can be seen in Appendix B, with the results for the Filler--Gap dependency tests for the GRNN model in the right facet of Figure \ref{fig:resid_analysis}.\footnote{With the MVRR test suite, no conditions are technically ungrammatical, however we treat the \textit{reduced\_ambiguous} condition as ungrammatical for the purposes of this analysis.} Across all tests, we find that ungrammatical conditions show much higher residual error. The mean absolute value of the residual error is 163$ms$ in grammatical conditions, but in ungrammatical conditions it is 244$ms$. The values of the two conditions are significantly different ($p<0.001$ by a $t$-test). Generally, residuals are largest for Cleft, Filler--Gap Dependency and MVRR suites, and smaller for suites that involve NPI Licensing, Anaphora agreement and Subject-Verb Number agreement. Human reaction-times are known to be susceptible to interference effects from distractors for these syntactic phenomena \cite{jager-etal:2020-interference-patterns}, which may explain why residuals are smaller for these suites. Taken together this analysis demonstrates that model surprisal values specifically under predict human reaction times in ungrammatical critical regions, suggesting that they are less sensitive to syntactic violations than are humans.

\section{Discussion}

Our experiments have tackled the question of whether syntactic difficulty can be reduced to by-word probabilities by providing a comparison of Language Model and human behavior that is both incremental and targeted. Our methods build on those presented in \citet{van2018modeling} and \citet{van-schijndel_linzen_2020}, but differ from theirs in a number of key respects, which we review briefly below to highlight to novel aspects of our own investigation. First, all of our test suites target grammatical/ungrammatical contrasts (except for the MVRR gardenpath test), whereas \citeauthor{van-schijndel_linzen_2020} test locally ambiguous sentence regions that (may) require re-analysis for proper processing. Second, we assess a broad range of grammatical violations across sixteen test suites that target seven distinct structures. Third, we deploy a novel measurement of processing time (\textit{Interpolated Maze}), instead of self-paced reading. We fit our own linear models from the I-Maze data, and use a ms/bit scalar term derived from lexical distractor items. Finally, we provide a novel analysis that compares the residuals of linear fits between critical and non-critical regions, and we break down these residuals based on the grammaticality of the condition.

\subsection{Model Comparison}

\begin{figure*}
    \centering
    \includegraphics[width = 1\textwidth]{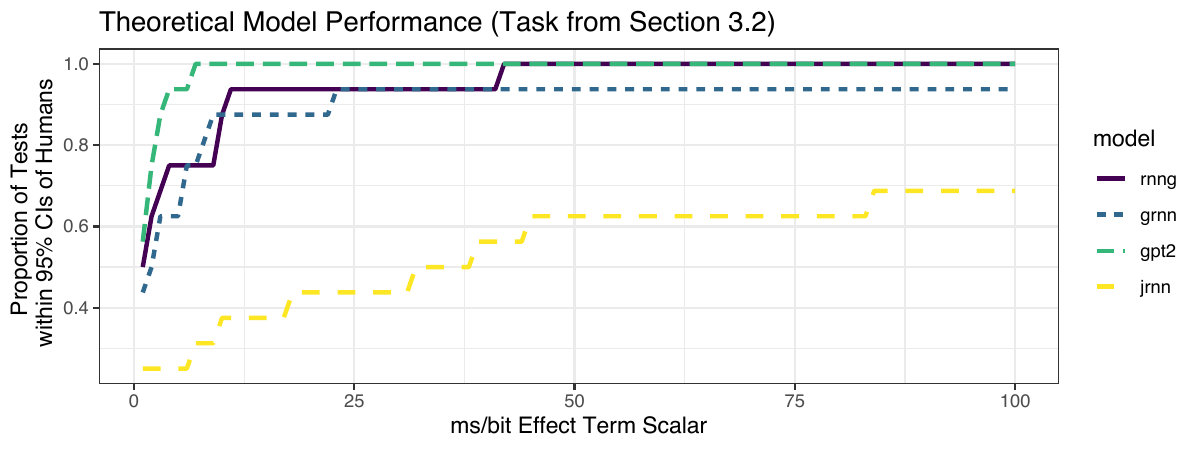}
    \caption{The effect of an additional ms/bit scalar term on model performance from tests in Section \ref{sec:slowdown}. Results indicate that both the RNNG and GRNN models could reach near human-like performance (within the human confidence intervals 90\% of the time) when the scalar term is around 10.}
    \label{fig:theoretical-performance}
\end{figure*}

While none of our models is able to capture humanlike sensitivity in ungrammatical critical regions, we do see some variation between them, with RNNG and GPT-2 in particular showing the most humanlike results. To compare model performance for accuracy scores (i.e. the results presented in Section \ref{sec:accuracy}), we fit pairwise logistic regression models, with the model class as the sole predictor, and random slopes for nested item/test suite combinations and predictions (this because predictions are shared across test suites of the same type). We find that GPT-2 performs significantly beter than both JRNN and GRNN ($p<0.01$) and the contrast between RNNG and GRNN approaches significance ($p=0.07$) None of the other pairwise comparisons are significant.


To compare model performance at predicting human slowdown in critical regions, we look at the difference in residual errors between the models from Section \ref{sec:residuals} in the critical regions. We fit liner regression models with the residual as predictor variable, nested item/test suite combinations, and \textit{condition} as random slopes. We find a significant contrast between GPT-2 and JRNN ($p<0.05$), with GPT-2 performing better, and a near-significant contrast between RNNG and JRNN ($p=0.053$). Overall, these results support the conclusion that GPT-2 and RNNG have a mild advantage over the other models. This is especially interesting for the RNNG model, given that it was trained on orders of magnitude less data than GPT-2.

\subsection{Single Stage Models}

For the last decade, a ``single-stage'' theory of incremental processing \citep{levy2008expectation}, in which word surprisal in a left-to-right language model (with a large or unlimited beam for models that explicitly represent multiple incremental parses) is the sole determinant of the processing difficulty that arises due to the relationship between a word and the context it appears in, has been a prominent candidate theory for both experimental \citep{staub:2011word-recognition} and computational \citep{frank-bod:2011} psycholinguistic investigations.  Although such a ``single-stage'' can capture the \emph{qualitative} difficulty patterns induced by garden-pathing and other grammar-based expectation violations \citep{hale2001probabilistic,levy:2013sentenceProcessing}, we now see that it \emph{quantitatively} under-predicts the difficulty induced when grammatical expectation violations are involved, as measured by self-paced reading \citep{van-schijndel_linzen_2020} and response times in the Maze task (here).

But just how bleak is the outlook for single-stage models? To investigate this, we re-analyze the results from Section \ref{sec:slowdown} with theoretical model performance that includes an additional scalar term that corresponds with an increase in the slope for surprisal relative to that obtained from the fit to reaction times. The results in Figure \ref{fig:theoretical-performance}. Here, the y-axis shows the proportion of tests for which the models are within the confidence intervals of human results, and the x-axis shows this scalar term. We find that models achieve 90\% accuracy levels when the scalar term is 4 for GPT2, 11 for RNNG and 23 for GRNN. What this means is that if either the ms/bit scalar term, or the surprisal in ungrammatical conditions were (slightly under) an order of magnitude greater, then the models' performance would match humans.

While we agree with the assessment from \citet{van-schijndel_linzen_2020} that these results pose a challenge for contemporary implemented models, we do not necessarily believe that they cannot be overcome within the framework of single-stage models, especially ones that are mediated by symbolic representations like the RNNG. Multiple options exist that could magnify surprisal values in locally ambiguous or ungrammatical regions, such as a reduced beam size \citep{roark2001probabilistic} or particle filters \citep{levy2009modeling}. Taken together, these recent results highlight a key question for future research---what additional modeling mechanisms will be needed to accurately predict not only qualitative but also quantitative patterns of human difficulty in language processing.

\section*{Acknowledgements}

RPL gratefully acknowledges NSF grant BCS-1551866, a Google Faculty Research Award, and funding from the MIT–IBM AI Lab.

\newpage
\section*{Ethical Considerations}

Data were collected under an Institutional Review Board (IRB) approved protocol for online human subject experimentation. Participants were compensated \$2.00 for their participation in I-Maze experiments. Experiments took $\sim$15 minutes, which meant participants were being compensated $\sim$\$8.00/hour. We chose this rate because it is slightly above federal minimum wage, which we take to be a fair baseline for compensation. All information associated with experimental participants was anonymized prior to analysis.

\bibliographystyle{acl_natbib}
\bibliography{acl2021}

\appendix

\section{Consistency/Accuracy Scores by Prediction}

\begin{figure}
    \centering
    \includegraphics[width=0.95\textwidth]{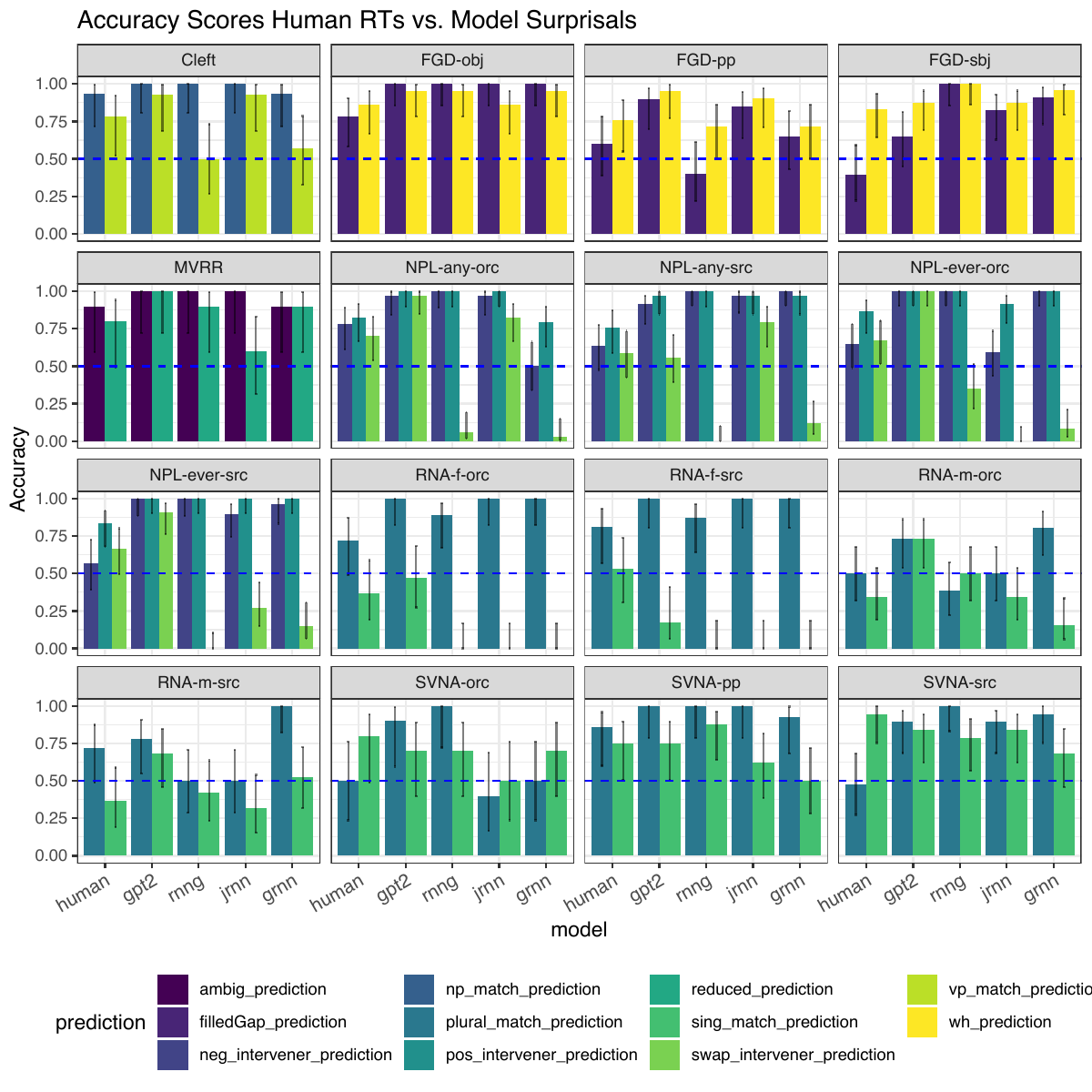}
    \caption{Test suite accuracy / consistency scores broken down by individual predictions.}
    \label{fig:acc_by_prediction}
\end{figure}

Figure \ref{fig:acc_by_prediction} gives accuracy scores for humans and LM models, broken down by individual predictions. Predictions are taken from \cite{hu2020systematic}, outlined in their Appendix B. Prediction names correspond to the licensed element of the sentence, so \textit{sing\_match\_prediction} for reflexive anaphora licensing corresponds to the contrast where \textit{himself} or \textit{herself} is grammatical (as opposed to \textit{themselves}). Accuracy/consistency scores are similar between humans and models for cleft structures, filler--gap dependencies (except for \textit{subject} tests, which we discuss below), MVRR gardenpath and Subject Verb Number Agreement suites. In the rest of this appendix, we focus in on structures that show different accuracy/consistency score patterns for humans and models.

For filler--gap dependency tests, the human data differs from the model data when there is a gap in the \textit{subject} position (FGD-sbj test). In this case, both achieve relatively high scores for the \textit{wh prediction} (yellow bars), but lower scores filled-gap prediction (\texttt{I know *who/that my mother...}). (It should be noted that this contrast is not one strictly of grammaticality in the critical region, as the sentence could be felicitously completed by a gap in the object position.) This behavior is in perfect alignment with the large amount of data demonstrating that English speakers take longer processing object gaps over subject gaps, and suggests that such expectations are weaker in our neural models.

Turning to NPI and anaphor licensing, we see a consistent pattern of difference between humans and models. For the NPI tests, models perform much worse than humans at the \textit{swap\_intervener} predictions (\texttt{No senator that the lawyer liked ... ever/any} vs. \texttt{The senator that no lawyer liked ... ever/any}), whereas human participants performed about as well on these tests as on the others. For reflexive anaphora licensing, human performance is worse for the \textit{singular} predictions, regardless of the gender of the pronoun, indicating a plural bias across the board. For models, this is true only for the feminine pronoun (\textit{herself}), and the difference in accuracy is much greater than the human difference in consistency scores. When the masculine version of the pronoun is used, models show similar scores for both the \textit{singular} and \textit{plural} predictions. This pattern is consistent with a plural bias in humans, but a bias against specifically the feminine (singular) form of the pronoun in models.

\section{Linear Fit Residuals by Condition}

\begin{figure*}[t]
    \centering
    \includegraphics[width = 0.99\textwidth]{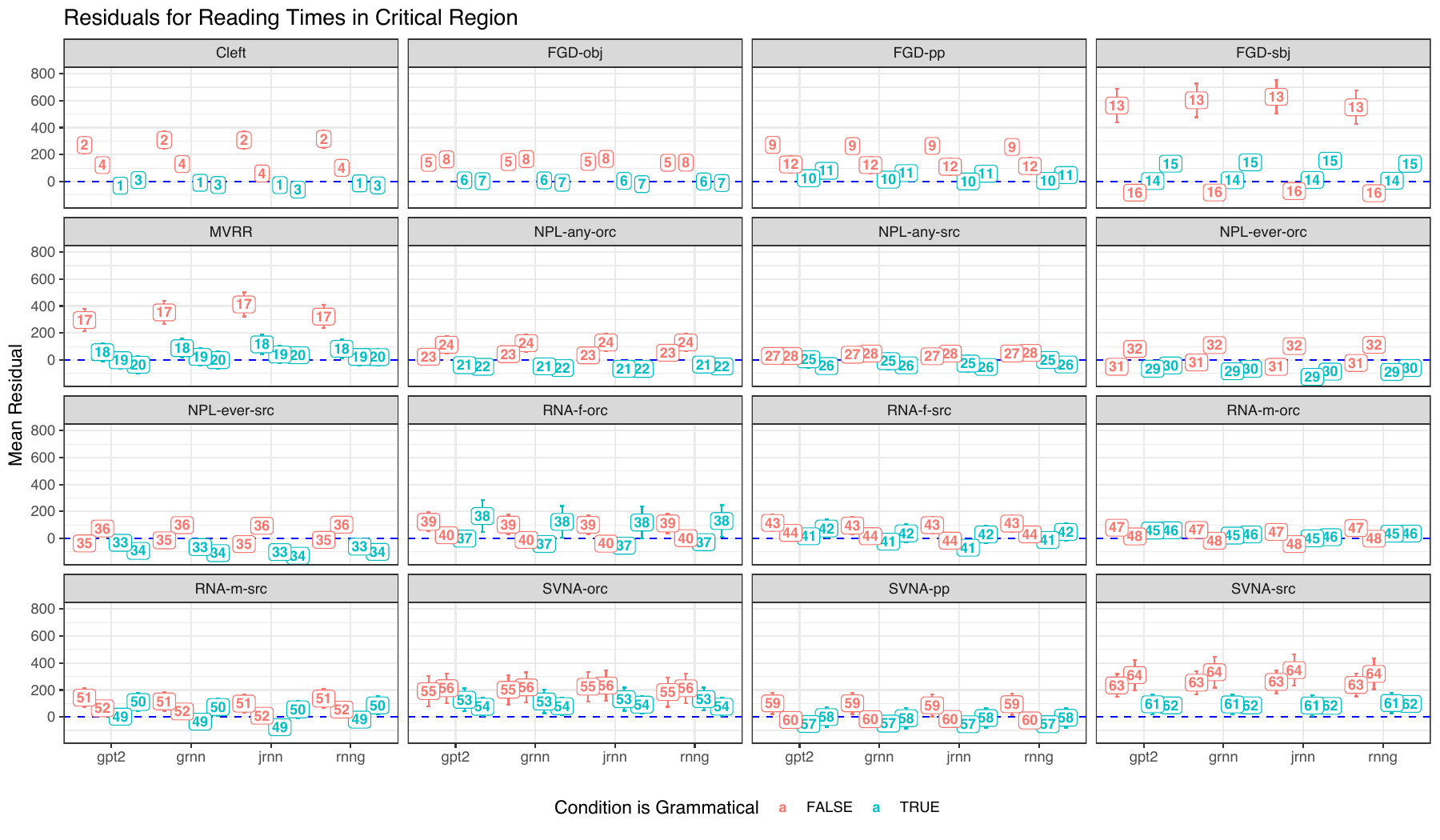}
    \caption{Residuals for predicted reaction times in critical regions, from a linear fit trained to predict reaction times from surprisal values in non-critical regions. Labels indicate condition name, with a reference provided in Appendix A. Error bars are 95\% confidence intervals. Across the majority of test suites, ungrammatical conditions show larger residuals, indicating that they are predicted less well by LM surprisal values.}
    \label{fig:resid_breakdown}
\end{figure*}

Table \ref{tab:conditions} gives a breakdown of all test suite conditions, with an example and a tag used for labeling for the left panel of Figure \ref{fig:resid_analysis} in the main text and for the figures in this appendix. Ungrammatical conditions are marked with a star. Figure \ref{fig:resid_breakdown} shows the residuals from our linear fits for each condition/test suite pair. See the figure caption for more detail.

\begin{table*}[!ht]
    \centering
    \footnotesize
    \resizebox{\textwidth}{!}{\begin{tabular}{llll}
        \toprule
        Condition Label & Test Suite Name & Condition Name & Example \\
        \midrule
        1&Cleft&np-match&What she spied was the giraffe\\
        2&Cleft&np-mismatch&*What she spied was see the giraffe\\
        3&Cleft&vp-match&What she did was see the giraffe\\
        4&Cleft&vp-mismatch&*What she did was see the giraffe\\
        \midrule
        5&FGD-obj&that-gap&*I know that my mother sent --- to Taylor yesterday.\\
        6&FGD-obj&that-nogap&I know that my mother sent the present to Taylor yesterday.\\
        7&FGD-obj&what-gap&I know what my mother sent --- to Taylor yesterday.\\
        8&FGD-obj&what-nogap&*I know what my mother sent the present to Taylor yesterday.\\
        9&FGD-pp&that-gap&*I know that my mother sent the present to -- yesterday.\\
        10&FGD-pp&that-nogap&I know that my mother sent the present to Taylor yesterday.\\
        11&FGD-pp&what-gap&I know who my mother sent the present to --- yesterday.\\
        12&FGD-pp&what-nogap&*I know who my mother sent the present to Taylor yesterday.\\
        13&FGD-sbj&that-gap&*I know that --- sent the present to Taylor yesterday.\\
        14&FGD-sbj&that-nogap&I know that my mother sent the present to Taylor yesterday.\\
        15&FGD-sbj&what-gap&I know who --- sent the present to Taylor yesterday.\\
        16&FGD-sbj&what-nogap&*I know who my mother sent the present to Taylor yesterday.\\
        \midrule
        17&MVRR&reduced-ambig&The woman painted a portrait was impressed by it.\\
        18&MVRR&reduced-unambig&The woman drawn a portrait was impressed by it.\\
        19&MVRR&unreduced-ambig&The woman who was painted a portrait was impressed by it.\\
        20&MVRR&unreduced-unambig&The woman who was drawn a portrait was impressed by it\\
        \midrule
        21&NPL-any-orc&neg-neg&No senator that no journalist likes has gotten any votes.\\
        22&NPL-any-orc&neg-pos&No senator that the journalist likes has gotten any votes.\\
        23&NPL-any-orc&pos-neg&*The senator that no journalist likes has gotten any votes.\\
        24&NPL-any-orc&pos-pos&*The senator that the journalist likes has gotten any votes.\\
        25&NPL-any-src&neg-neg&No senator that likes no journalists has gotten any votes.\\
        26&NPL-any-src&neg-pos&No senator that likes the journalists has gotten any votes.\\
        27&NPL-any-src&pos-neg&*The senator that likes no journalists has gotten any votes.\\
        28&NPL-any-src&pos-pos&*The senator that likes the journalist has gotten any votes.\\
        29&NPL-ever-orc&neg-neg&No senator that no journalist likes has ever won.\\
        30&NPL-ever-orc&neg-pos&No senator that the journalist likes has ever won.\\
        31&NPL-ever-orc&pos-neg&*The senator that no journalist likes has ever won.\\
        32&NPL-ever-orc&pos-pos&*The senator that the journalist likes has ever won.\\
        33&NPL-ever-src&neg-neg&No senator that likes no journalists has ever won.\\
        34&NPL-ever-src&neg-pos&No senator that likes the journalists has ever won.\\
        35&NPL-ever-src&pos-neg&*The senator that likes no journalists has ever won.\\
        36&NPL-ever-src&pos-pos&*The senator that likes the journalist has ever won.\\
        \midrule
        37&RNA-f-orc&match-plural&The queens who the dukes mistrust saw themselves in the mirror.\\
        38&RNA-f-orc&match-sing&The queen who the duke mistrusts saw herself in the mirror.\\
        39&RNA-f-orc&mismatch-plural&*The queens who the dukes mistrust saw herself in the mirror.\\
        40&RNA-f-orc&mismatch-sing&*The queen who the dukes mistrust saw themselves in the mirror.\\
        41&RNA-f-src&match-plural&The queens who hunted the rabbit saw themselves in the mirror.\\
        42&RNA-f-src&match-sing&The queen who hunted the rabbits saw herself in the mirror.\\
        43&RNA-f-src&mismatch-plural&*The queens who hunted the rabbit saw herself in the mirror.\\
        44&RNA-f-src&mismatch-sing&*The queen who hunted the rabbits saw themselves in the mirror.\\
        45&RNA-m-orc&match-plural&The dukes who the dukes mistrust saw themselves in the mirror.\\
        46&RNA-m-orc&match-sing&The duke who the duke mistrusts saw himself in the mirror.\\
        47&RNA-m-orc&mismatch-plural&*The dukes who the dukes mistrust saw himself in the mirror.\\
        48&RNA-m-orc&mismatch-sing&*The duke who the dukes mistrust saw themselves in the mirror.\\
        49&RNA-m-src&match-plural&The dukes who hunted the rabbit saw themselves in the mirror.\\
        50&RNA-m-src&match-sing&The duke who hunted the rabbits saw himself in the mirror.\\
        51&RNA-m-src&mismatch-plural&*The dukes who hunted the rabbit saw himself in the mirror.\\
        52&RNA-m-src&mismatch-sing&*The duke who hunted the rabbits saw themselves in the mirror.\\
        \midrule
        53&SVNA-orc&match-plural&The lawyers that helped the mayor are organized.\\
        54&SVNA-orc&match-sing&The lawyer that helped the mayors is organized.\\
        55&SVNA-orc&mismatch-plural&*The lawyers that helped the mayor is organized.\\
        56&SVNA-orc&mismatch-sing&*The lawyer that helped the mayors are organized.\\
        57&SVNA-pp&match-plural&The lawyers that the mayor helped are organized.\\
        58&SVNA-pp&match-sing&The lawyer that the mayors helped is organized.\\
        59&SVNA-pp&mismatch-plural&*The lawyers that the mayor helped is organized.\\
        60&SVNA-pp&mismatch-sing&*The lawyer that the mayors helped are organized.\\
        61&SVNA-src&match-plural&The lawyers next to the mayor are organized.\\
        62&SVNA-src&match-sing&The lawyer next to the mayors is organized.\\
        63&SVNA-src&mismatch-plural&*The lawyers next to the mayor is organized.\\
        64&SVNA-src&mismatch-sing&*The lawyer next to the mayors is organized.\\
        \bottomrule
    \end{tabular}}
    \caption{Conditions for each of the test suites assessed in this paper, with a tag (used for labeling in Figure \ref{fig:resid_breakdown}) and an example. Ungrammatical sentences are marked with a star ($*$)}
    \label{tab:conditions}
\end{table*}

\end{document}